\documentclass[letterpaper, 10 pt, conference]{ieeeconf}  % Comment this line out if you need a4paper
\usepackage{caption}
\usepackage{subcaption}

\usepackage{todonotes}

\IEEEoverridecommandlockouts                              % This command is only needed if 
\usepackage{graphicx}
\usepackage{amsmath} % assumes amsmath package installed
\usepackage{listings}

\title{\LARGE \bf
Physical Neural Cellular Automata for 2D Shape Classification}

\author{%XX$^{1}$, XX$^{1}$, XX$^{1}$, XX$^{1}$, XX$^{1}$, XX$^{1}$
Kathryn Walker, Rasmus Berg Palm, Rodrigo Moreno, Andres Faina, Kasper Stoy, Sebastian Risi\\
IT University of Copenhagen\\
\{kwal, rasmb, rodr, anfv, ksty, sebr\}@itu.dk
}
\begin{document}

\maketitle
\thispagestyle{empty}
\pagestyle{empty}

%%%%%%%%%%%%%%%%%%%%%%%%%%%%%%%%%%%%%%%%%%%%%%%%%%%%%%%%%%%%%%%%%%%%%%%%%%%%%%%%
\begin{abstract}

Materials with the ability to self-classify their own shape have the potential to advance a wide range of engineering applications and industries. Biological systems possess the ability not only to 
self-reconfigure but also to self-classify themselves to determine a general  
shape and function.
Previous work into modular robotics systems has only enabled 
self-recognition and self-reconfiguration into a specific target shape, missing the inherent robustness present in nature to self-classify. In this paper we therefore take advantage of recent advances in deep learning and neural cellular automata, and present a simple modular 2D robotic system that can infer its own class of shape through the local communication of its components. Furthermore, we show that our system can be successfully transferred to hardware which thus opens  opportunities for future self-classifying machines. Code available at https://github.com/kattwalker/projectcube. Video available at https://youtu.be/0TCOkE4keyc.
%TODO: include for final version
%All videos associated with this project can be viewed here: https://youtu.be/VpbHqw58vPM . Code and PCB design can be found here: https://github.com/kattwalker/projectcube.git.

%Materials with the ability to self-recognize and self-classify their own class of shape and remodel in response
%to environment and or damage have the potential to advance a wide range of engineering applications
%and industries. Whilst such mechanisms are widely used by biological systems, e.g.\ 
%to grow a target shape correctly or to remodel body parts (e.g.\ the tail of a
%salamander regrows after being removed), they have not yet been exploited in
%machine design. Taking advantage of recent advances in deep learning and neural cellular automata, we present a simple modular 2D robotic system that can infer its own class of shape only through the local communication of its components. All videos associated with this project can be viewed here: https://youtu.be/VpbHqw58vPM

%the aim ofthis proposal is to create a large-scale physical system capable of self-shape recognition only through the local communication of thousands of identical cells (modular cubes) that do not need to be aware of their 3D position. We anticipate that the technologies developed in this proposal could not only enable more robust robots but in the future also e.g. buildings made from self-modeling materials that cancontinuously check for weaknesses in their structural integrity.

\end{abstract}

%%%%%%%%%%%%%%%%%%%%%%%%%%%%%%%%%%%%%%%%%%%%%%%%%%%%%%%%%%%%%%%%%%%%%%%%%%%%%%%%
\section{INTRODUCTION}

Many biological systems have the remarkable ability to correctly determine their anatomical structure.
For example, just through the process of local communication and self-organization, cell groups can
determine whether they formed a certain target shape correctly (e.g.\ an organ). Furthermore, they can even remodel body parts after damage. For example, the tail of a salamander can regrow and remodel into a leg after damage \cite{farinella1956transformation,vieira2020advancements}, and simple organisms such as Hydra and Planaria are capable of complete morphological repair, regardless of which body part is removed \cite{levin2019endogenous,vogg2019model}.  
Their ability to self-classify general anatomy, rather than self-recognise one single target shape allows for variance between individuals, thus making the entire process more robust. For instance, the overall function and design of an organ may be the same, but the final specific shape, size or scale between different individuals of a species will vary.

%Many biological systems have the remarkable ability to correctly determine their anatomical structure, even when the structure consists of many thousands/millions of cells. For example, just through the process of local communication, self-organization and self-classification, cell groups can determine whether they formed a certain class of target shape correctly (e.g.\ an organ) \cite{joshi2012epithelial} or if a specific body part is in need of repair and remodeling (e.g.\ the tail of a salamander can regrow and remodel into a leg after being removed \cite{farinella1956transformation}).

Artificial engineering systems formed of thousands of individual modules with the ability to infer their own class of shape  
%and then to potentially remodel parts to a form a different class of robot
could be desirable in many applications. This ability could even enable artificial systems to go beyond nature in terms of morphological adaption to damage or to new environments by completely re-configuring their body shape. Led by the promise of versatile robotic systems, the field of modular robotics, 
%i.e.\ multiple robots that are able to communicate with one another to form an overall larger complex shape, 
has existed for almost 30 years. Whilst many of these systems are able to self-recognise and self-reconfigure into a specific target shape, they are missing the inherent robustness present in nature to self-classify \cite{thalamy2019survey}. Furthermore, much of the modular robot self-recognition work exists only in simulation and is not proven to work in hardware, limiting its usefulness for real world  implementations and applications.

%Artificial engineering systems formed of thousands of individual modules with the ability to infer their own class of shape and then to potentially remodel parts to a form a different class of robot could be desirable in many applications. Led by the promise of versatile robotic systems that are able to adapt to new situations, tasks and environments potentially better than their traditional robotic counterparts, the field of modular robotics has existed for almost 30 years.

In this work, we build on the previous contributions from the world of modular robotics, deep learning and neural cellular automata \cite{mordvintsev2020growing,randazzo2020self}. Neural cellular automata are a derivative of the simpler cellular automata and consist of a regular grid of cells where each cell can be in any one of a finite set of states. Individual cell states are then updated based on information from their neighbour's states and simple rule sets. In neural cellular automata, these simple rules are replaced by trained artificial neural networks.

We investigate how methods for shape classification through neural cellular automata \cite{randazzo2020self}, which have so far only been explored in perfect simulated environments, can be extended to work in hardware. 
%Our work is an extension of previous examples such as \cite{mordvintsev2020growing,randazzo2020self},  further discussed in the related work section, where neural cellular automata are able to self-classify different shapes formed of hundreds of individual cells. 
 %Previous work have all been completed in a perfect simulated environment and not transferred to hardware. 
 Whilst the usefulness of a robust physical system is undeniable, so are the reality gap challenges of moving from  simulation to hardware, particularly when dealing with multi-robot  systems \cite{dorigo2021swarm}.
% with the obvious prevalence of the reality gap \cite{dorigo2021swarm}. 

Therefore, the main contribution of our work is the design and implementation of a physical version of neural cellular automata capable of self-classifying 10 distinct classes of shape. Our work is able to use sparse representation to classify the overall shape of the cells, which is particularly relevant in a hardware setting where communication limitations might occur. That is, the whole ``map" of the robot does not need to be communicated between every module. 
%and no specific target shape required.
This way, the system also displays some inherent scale invariance when classifying differently sized shapes, even though it was not trained for it. Increasing this type of robustness is an important open challenge in modular robotics \cite{thalamy2019survey}.
%This means that our system is also able to inherently classify the different shapes regardless of size/scale, also allowing for robustness in the system, even though this was not trained for, which is cited as an open challenge in modular robotics \cite{thalamy2019survey}.    

\section{RELATED WORK}

The work detailed here is closely related to the field of modular robotic self assembly, i.e.\ multiple robots that are able to communicate with one another to form an overall larger complex shape \cite{thalamy2019survey}. %, both in the modular robotics and swarm domains (sometimes referred to as mobile modular robot~\cite{thalamy2019survey}).
Modular robotics systems were first introduced over 30 years ago by Fukuda et al. \cite{fukuda1990cellular,fukuda1994cellular}. Unsurprisingly, since then, they have been researched in great detail. These modular systems hold the promise to generate versatile robots that are able to adapt to new situations, tasks and environments potentially better than their traditional counterparts.

Some attempts at designing re-configuring modular robots use centralised algorithms \cite{kotay2000algorithms,baca2017configuration,bie2017approach}. In these approaches, each module sends its own state to an external agent, which then uses all the information gathered to build a map of the current robot configuration. Clearly this method relies on the external agent for success. 

Other researchers have focused on decentralised control, where each module is responsible for its own behaviour, and state changes are made based on local information gathered from neighbours and simple rules \cite{thalamy2019survey}. This approach is closely aligned to the concept of cellular automata in the work presented here. Usually, each module is given a specific target overall shape and its deferred local state includes some encoding of global (albeit sometimes partial) information gathered from neighbours. Note, in some cases emergent modular robot shapes have also been investigated through simple local rule interactions in decentralised agents \cite{slavkov2018morphogenesis}. 

Using a variety of different techniques and algorithms (see Thalamy et al.~\cite{thalamy2019survey} for a comprehensive survey) this active research community has achieved excellent results. For example, a pioneering example is the work by Rubenstein et al.~\cite{rubenstein2014programmable} which utilized a 1024 robot swarm~\cite{rubenstein2012kilobot}. In this decentralised approach, each agent was programmed with a target shape and the algorithm required to determine their necessary position within the shape. A similar swarm approach has also been carried out in 3D~\cite{stoy2004self,stoy2006using} using a target CAD model, although this has only been proven to work in simulation. Other examples of work proposing different methods for modular robot configuration exist~\cite{meng2011autonomous,sprowitz2010roombots,murata1994self,mathews2017mergeable} but are also carried out mainly in a purely simulated environment. The amount  of systems existing in hardware is less, although some do exist for minimal amounts of modules  \cite{salemi2006autonomous,liu2020configuration,sprowitz2014roombots,yim2001distributed}.

All the above examples (both centralised and decentralised) concentrate on algorithms to create one explicit predefined shape and are therefore not able to differentiate between different classes of shape. If, for example, the modules are required to configure into a chair, they require a specific target chair shape (e.g.\  a single CAD model of a chair \cite{stoy2004self}). Therefore, if the modules are unable to form that specific shape, they have failed, regardless of whether the actual shape formed meets the required specification/function. There is an inherent robustness missing in these systems;  they are unable to classify the different \textit{classes of shapes} regardless of size/scale/small changes in the overall design. As these systems are only programmed for one shape, they have no concept of their new configuration if the shape is damaged or otherwise changed through external influence. Although many of the current example would be able to recalculate their new shape, crucially they would not possess the knowledge as to whether or not this is a problem for the function of the system. They  can only self-recognise, not self-classify.

One system that has used a form of modular cells for self classification is  that by Randazzo et al. \cite{randazzo2020self} and our work most closely aligns here. In their work, shape classification based solely on the local communication of cells (decentralised, neural cellular automata) was shown to be possible in a simulated environment \cite{randazzo2020self}. The system is able to recognize MNIST digits (handwritten numbers), even with variation in the digit's shape or size. Whilst this technique, and the combination of collective intelligence with deep learning in general \cite{risi2021selfassemblingAI,ha2021collective}, can add inherent robustness to modular robotics, it has only been shown to work in the perfect conditions of a simulated environment.

Therefore, in the work presented here, we extend the concept of robust self-classifying neural cellular automata and design a hardware tile based system. Our system is capable of self classifying 10 distinct classes of shape (i.e., the number from 0-9) and has proven robustness to changes in scale.

\section{NEURAL NETWORK MODEL}

Our approach is based on Neural Cellular Automata (NCA) \cite{mordvintsev2020growing, wulff1992learning, nichele2017neat}. A Cellular Automata (CA) is a system in which ``cells" in a grid update their states according to the neighboring cells states and a set of rules. A prominent examples of a CA is Conway's game of life, which is Turing complete with binary states and just four simple rules. The NCA is a CA, in which the state of a cell is represented with a real valued vector, and the rules are functions parameterized by a neural network.

\begin{figure}[htbp!]
    \centering
    \includegraphics[width = 0.45\textwidth]{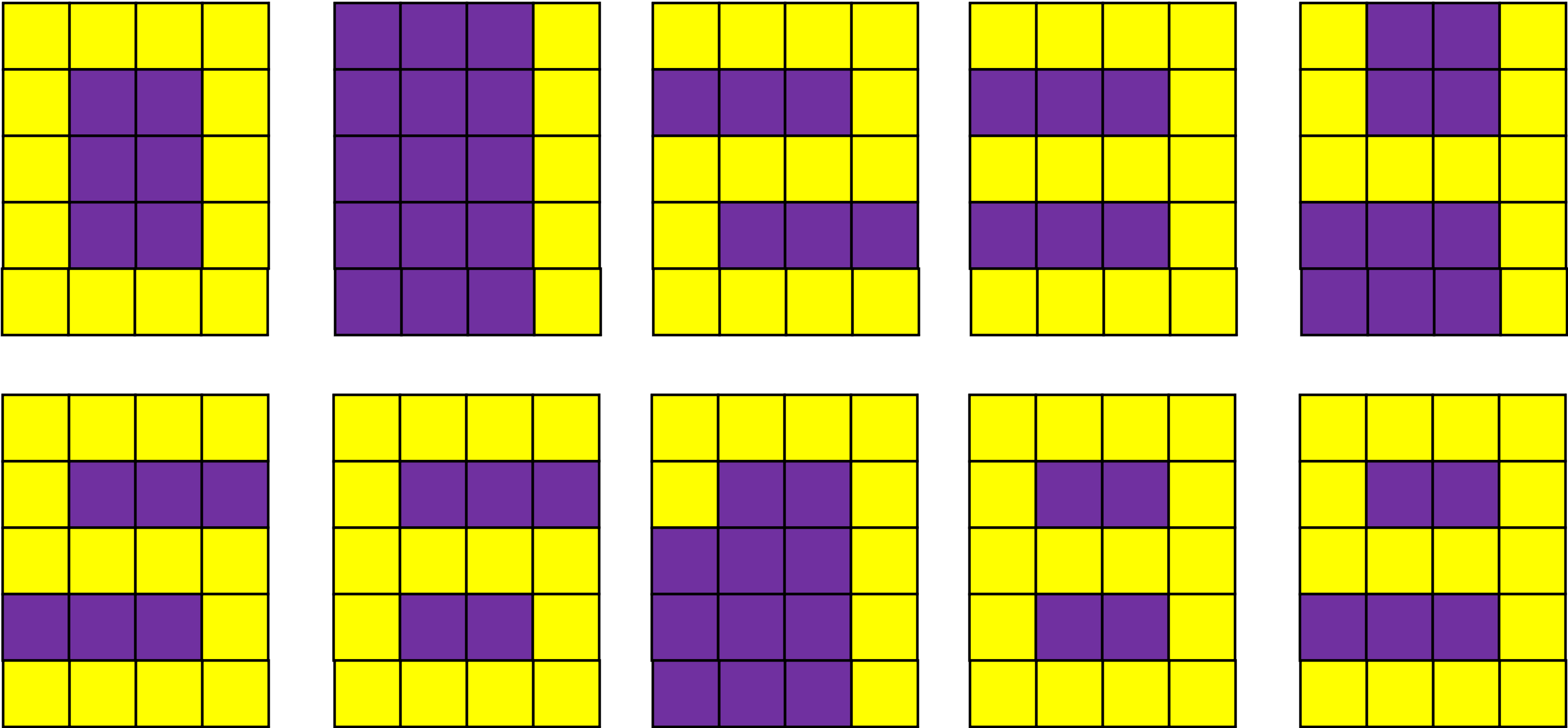}
    \caption{The configuration of cells for each of the 10 numbers (shapes). A yellow square represents an active cells, a purple square is an empty space on the grid.}
    \label{fig:training numbers}
\end{figure}

Building on previous NCA approaches \cite{randazzo2020self}, a  single trained NCA present in each cell, uses information gathered from its neighbours to classify the overall shape based on local communication. The system runs for a set number of steps and after each step, every cell outputs  a guess of what shape it believes it is a part of. In each cell the trained weights of the NCA are   the same, the difference is the local information gathered from neighbours that is the input for the NCA. Using the trained NCA,  after the set number of update each cell would successfully classify the overall shape. In our experiments, the cells are supposed to come to an agreement, which one of the ten different digits (Figure~\ref{fig:training numbers}) the overall shape represents.  

The vector valued state of cell $i$ at step $t$ is updated iteratively such that,
\begin{align}
    s_i^t = s_i^{t-1} + f_\theta(\{s_j^{t-1}\}_{j \in N(i)}) \,,
\end{align}
where $f_\theta$ is a neural network computing an additive update and $N(i)$ returns the indices of the neighbors of cell $i$ including $i$ itself. Here we use a state vector size of 21; in preliminary experiments we saw that smaller sizes did not work as reliably. All the cell states are initialized to a vector of zeros, except the first channel which is set to one.

In the case where the cells are aligned in a grid, and the neighbors are the $3 \times 3$ surrounding cells, $f_\theta$ can be efficiently implemented using a convolutional neural network \cite{gilpin2019cellular}. For our experiments we use a $3 \times 3$ convolution with 40 channels and a Rectified Linear Unit (ReLU) activation function \cite{nair2010rectified}, followed by a single $1 \times 1$ convolution also with 40 channels and relu, followed by a linear $1\times1$ convolution with 21 channels to match the cell state size. Note that the size of the neural network was kept conservatively small to ensure it would fit on the selected hardware micro-controller.

At any step the output of the NCA is defined as the argmax over the last $C$ channels, where $C$ is the amount of classes. We train the NCA to determine which one of ten different digits ($C=10$) its shape represents 
% to classify its own morphology into the $C=10$ shapes 
(Figure~\ref{fig:training numbers}), by minimizing the squared error between the last $C$ channels and the one-hot encoded class label $c_i$,
%s
\begin{align}
    \mathcal{L} =\sum_{i=1}^N (s_i^{T}[-C:] - c_i)^2
\end{align}
where $N$ is the amount of cells, $[-C:]$ indicates the last $C$ channels and $T$ is a random number of steps. For our experiments $T$ is uniformly sampled between $9$ and $29$. It would be more natural to use the softmax cross entropy loss for classification, but we found the squared error to be more stable. The loss is minimized with stochastic gradient descent for 2500 iterations, using a batch size of 128, and the Adam optimizer with the default parameters \cite{kingma2014adam}. The computations are done using Tensorflow \cite{abadi2016tensorflow}.

We modify the standard NCA setup described above to work in hardware in the following ways:

\textbf{Zero out kernel corners.} The hardware cells can only communicate with their neighbors in the four cardinal directions and not diagonally. To simulate this we clamp the diagonal weights of the $3\times3$ convolutional kernels to be zero, eliminating any diagonal communication while still allowing  efficient convolutional operations.

\textbf{Zero out empty grid cells.} The hardware cells can only communicate with other hardware cells, and not through the air. However, the convolutional neural network operates on the entire grid, and thus computes updates for the empty grid cells as well. To address this we clamp the empty grid cells to have an all zero state.

\textbf{Drop updates randomly.} We simulate the asynchronous nature of the hardware computation by dropping some updates at random during training. This is achieved by multiplying the computed updates by a random binary mask at each step. We randomly drop half of the updates at each step in this way.

\textbf{Validate with asynchronous simulation.} We validate the trained neural network in a simple asynchronous hardware simulation. At each step it samples the evaluation sequence of the $N$ cells \emph{with replacement}, such that 1) the evaluation order is random and 2) some cells may be evaluated more than once, and some not at all. See listing \ref{list:hwsim} for pseudo-code.

\begin{lstlisting}[language=python, basicstyle=\ttfamily\small, caption=Asynchronous hardware simulator, label={list:hwsim}, captionpos=b]
for i in range(n_steps):
  for j in range(len(cells)):
    random.choice(cells).evaluate()
\end{lstlisting}

\section{HARDWARE DESIGN}

Our hardware tiles are designed as a Printed Circuit Board (PCB) that sits on top the Arduino Mega 2560 (8 bit programmable prototyping board) as a shield. This provides a rigid body and allows us to simplify the electronics design of the module taking advantage of the 3 UART ports in the Arduino board for local communication among neighbor tiles. Table~\ref{tab:properties} summarises the main properties of the tile.

\begin{table}[htbp!]
\caption{Tile properties}\label{tab:properties}
\begin{tabular}{ll}
\textbf{Property}        & \textbf{Value}    \\ \hline
Base platform            & Arduino Mega 2560 \\
Dimensions               & 114.3x114.3 mm    \\
Connector current rating & 2 A               \\
Input voltage            & 12 V              \\
Communication            & Serial UART      
\end{tabular}
\end{table}
%* Takes advantage of the 3 existing UART ports plus a software serial port
%* Connectors are normal pin headers rated at 1A per pin and used double pins for power on each side
The shield is shaped as a square (114.3$\times$114.3 mm) with connectors in the middle of its four sides marked as the four cardinal directions.
Connectors are made of 2.54 mm pin headers and housings and are organized such that mating connectors (two male, two female) face each other when tiles are in the same orientation. The connector layout forces a specific order when connecting tiles. Nevertheless, pin headers provide a stable mechanical and electrical connection to neighboring tiles. 

Each connector has 6 pins, four of which transmit power, rated for up to 2A at 12 V, and allow the tiles to be powered with only one of them connected to a power source. Tiles can still be powered independently from each other by disconnecting the Arduino power source from the shield through dedicated jumpers. The other 2 pins connect to the TTL TX and RX pins of each of the UART ports on the Arduino and are organized so that connectors match their UART pins correctly with their mating connector. Since the Arduino only has 3 hardware UART peripherals,  the remaining UART connection is supplied using a software defined UART. 

%* Indicators include an RGB led and a 7 segment display
On the top side of the tiles, a RGB LED and a 7-segment display driven by the Arduino board, are used to indicate the number returned by the neural network. Four extra LEDs and a tactile switch can also be used to debug the tiles programming. Tiles are  programmed one by one with the current version of the hardware.

\begin{figure}[htbp!]
    \centering
    \includegraphics[width=0.4\textwidth]{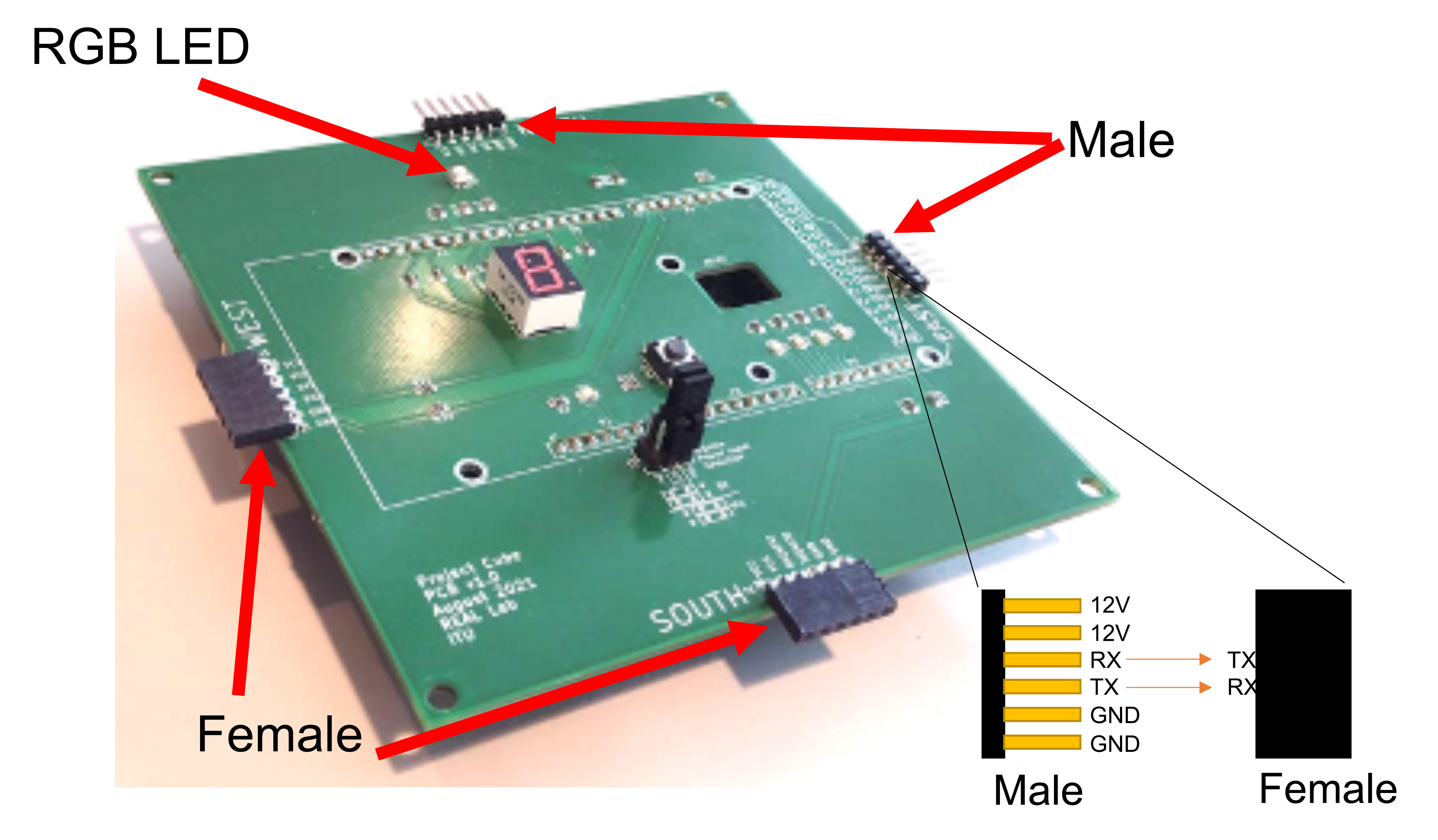}
    \caption{Left: PCB of tile top view. Right:  Connector pinout for the east side of the tile.}
    \label{fig:top_cube}
\end{figure}

%%%%%%%%%%%%%%%%%%%%%%%%%%%%%%%%%%%%%%%%%%%%%%%%%%%%%%%%%%%%%%%%%%%%%%%%%%%%%%%%
\section{Tile Firmware}%FIRMWARE/EXPERIMENT DESIGN}

Our system is designed so that all training of the neural network is carried out offline in simulation. 
%(as designed above in the Software Design section). 
This setup reduces the requirements of the firmware; the weights of the trained neural network are simply stored as a fixed array in the flash memory of the micro-controller. The  neural network occupies 18\% of the flash memory.

As all training of the neural network is done offline, the firmware programmed onto the tiles is relatively simple, consisting of three main functions; \textit{sending cell state}, \textit{receiving neighbouring cell states} and \textit{updating cell state}. The overall way these functions fit together is shown in Listing \ref{list:firm}. Each cell  updates after a fixed amount of time, regardless of whether it has received information from all its neighbours. For our experiments this time is set at the arbitrary value of 2 seconds. While the system can be run faster, this setting allowed us to more easily debug it. 

The number of cell updates is kept at 30, which is consistent with the neural network training. The number of directions is 4 (i.e., the 4 cardinal neighbours of the tile), however this can easily be scaled to 6 neighbours if the system is expanded into 3D.  

\begin{lstlisting}[language=c++, basicstyle=\ttfamily\small, caption=Module Firmware Overview, label={list:firm}, captionpos=b]
while(UPDATE_NUM < 30){
    for(int i = 0; i < NUM_DIRECTIONS; i++) {
             sendMessage();
          }
    for(int i = 0; i < NUM_DIRECTIONS; i++) {
             receiveMessage();
          }
    uint32_t now = millis();
    static uint32_t prev = 0;
    if((now - prev) > UPDATE_TIMEOUT_MS) {
              updateNeuralNet();
              UPDATE_NUM = UPDATE_NUM +1;
              prev = now;
            }
}
\end{lstlisting}

\textbf{Sending/Receiving Data.} The sending and receiving of cell states between neighbours occurs via the TTL TX and RX pins of each of the UART ports on the Arduino. Although for the majority of the program these cell states are stored as floats, during the communication phase they are converted by a linear scale to integer values between 0 - 255 to allow for communication via serial link. Although this introduces some noise to the system, it does not appear to be a problem, as suggested by the results shown below. 

\textbf{Updating Cell State.} To update the cell state, only a neural network forward pass is required as training takes place offline. The basic method is shown in Listing \ref{list:update}, where $n$, $e$, $s$, and $w$ are the cell states received from the north, east, south and west neighbours respectively. Note that if no neighbour is present, or the modules are out of sync and the neighbour has not yet sent a message, this data will be {0}.

\begin{lstlisting}[language=c++, basicstyle=\ttfamily\small, caption=Cell Update Overview, label={list:update}, captionpos=b]

x = relu(cell_state @ perceive_kernel[1,1] +
        n @ perceive_kernel[0,1] +
        e @ perceive_kernel[1,2] +
        s @ perceive_kernel[2,1] +
        w @ perceive_kernel[1,0] + 
        self.perceive_bias)

x = relu(x @ dmodel_kernel_1[0,0] 
    + dmodel_bias_1) 
x = x @ self.dmodel_kernel_2[0,0] 
    + dmodel_bias_2
cell_state = cell_state + x  
        
\end{lstlisting}

\section{RESULTS}

%The system was able to transfer successfully to hardware for all the trained shapes. 
In this section, we show the results for three of the shapes in detail, both for the simulated version and the hardware transfer. These results are consistent with the other shapes that the neural network is trained for.

\begin{figure}
     \centering
     \begin{subfigure}[b]{0.3\textwidth}
         \centering
         \includegraphics[width=\textwidth]{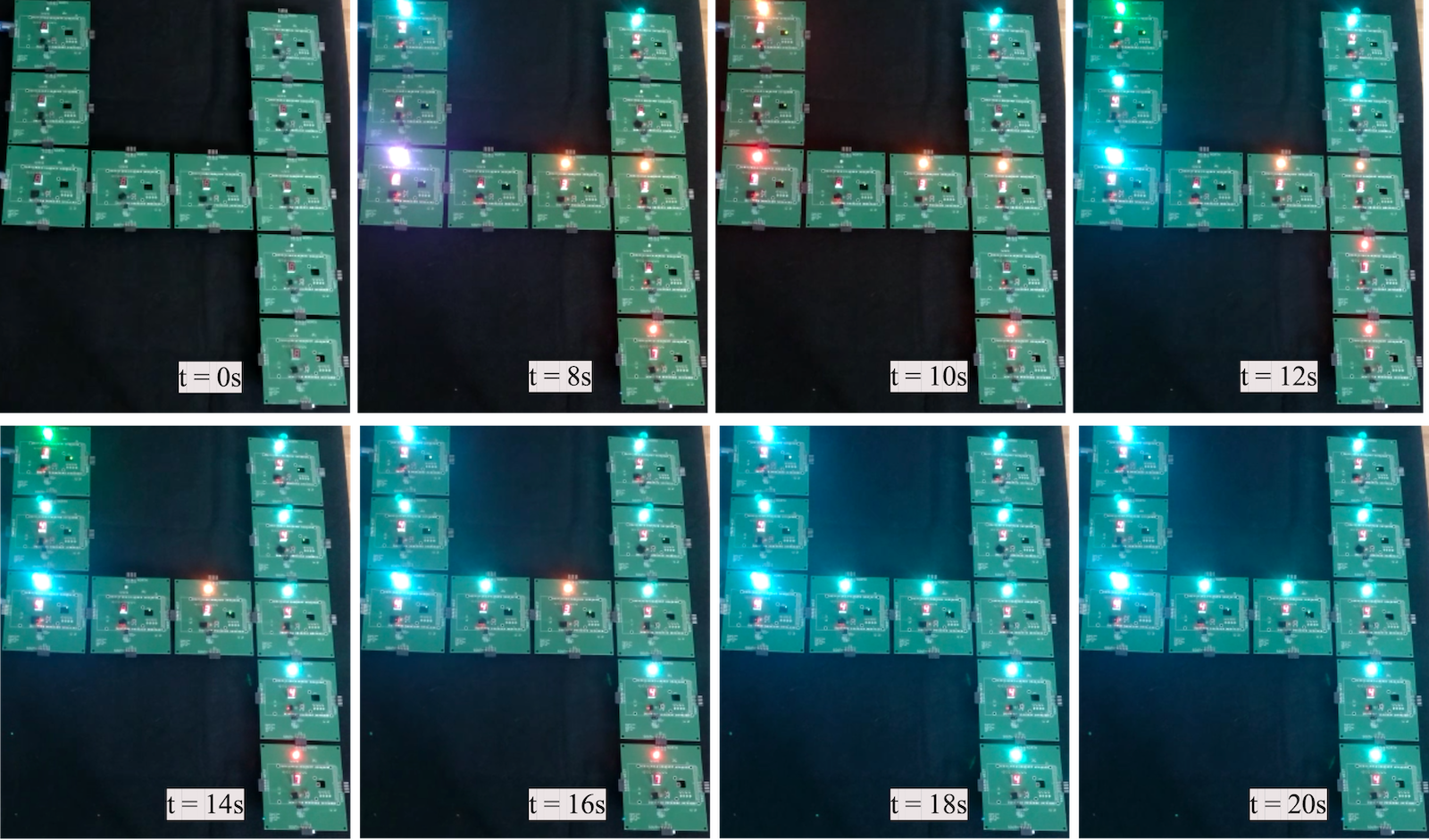}
         \caption{Results from hardware set-up of ``4".}
     \end{subfigure}
     \hfill
     \begin{subfigure}[b]{0.3\textwidth}
         \centering
         \includegraphics[width=\textwidth]{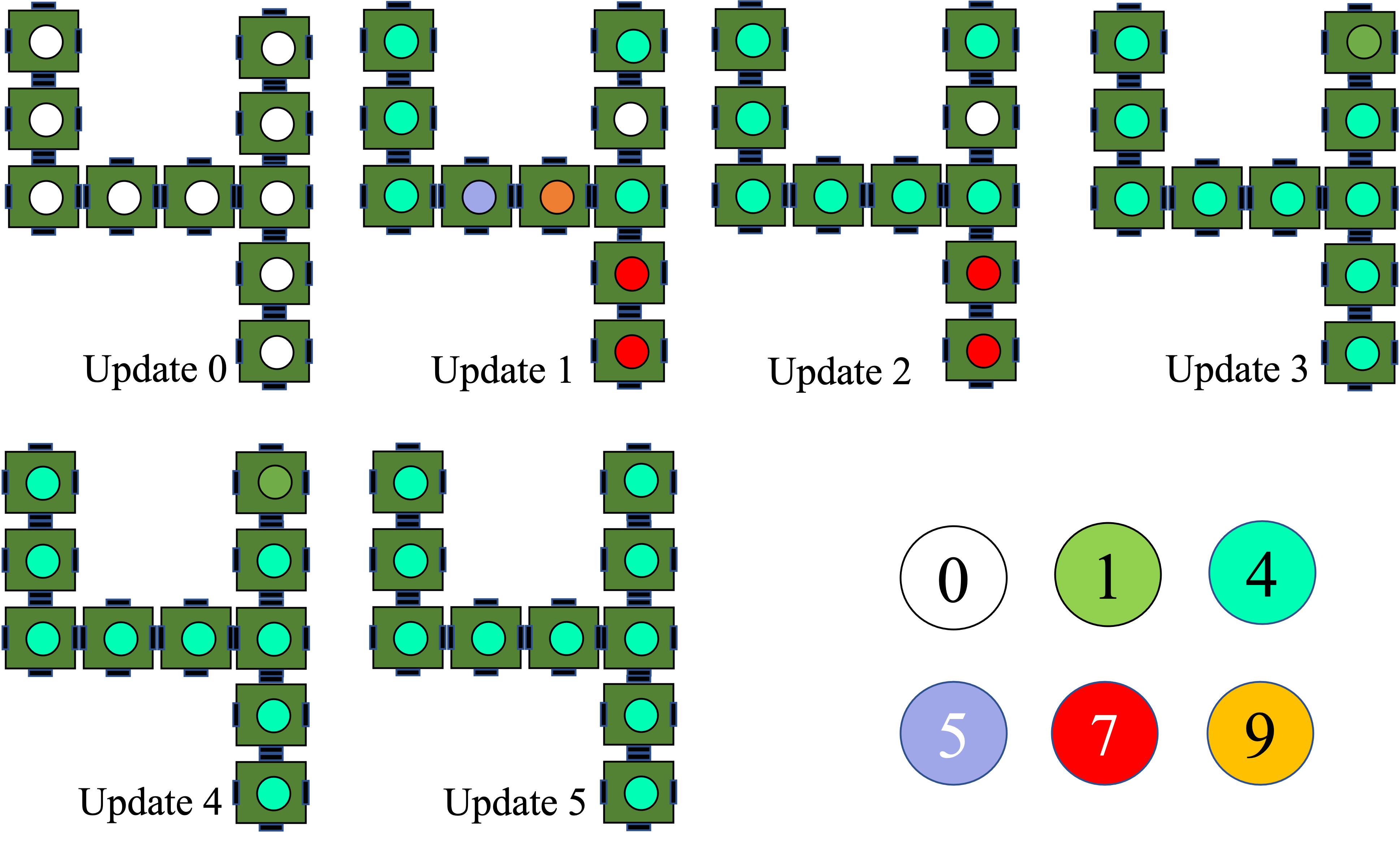}
         \caption{Results from the simulation of the ``4" setup.}
     \end{subfigure}
    \hfill
     \begin{subfigure}[b]{0.3\textwidth}
         \centering
         \includegraphics[width=\textwidth]{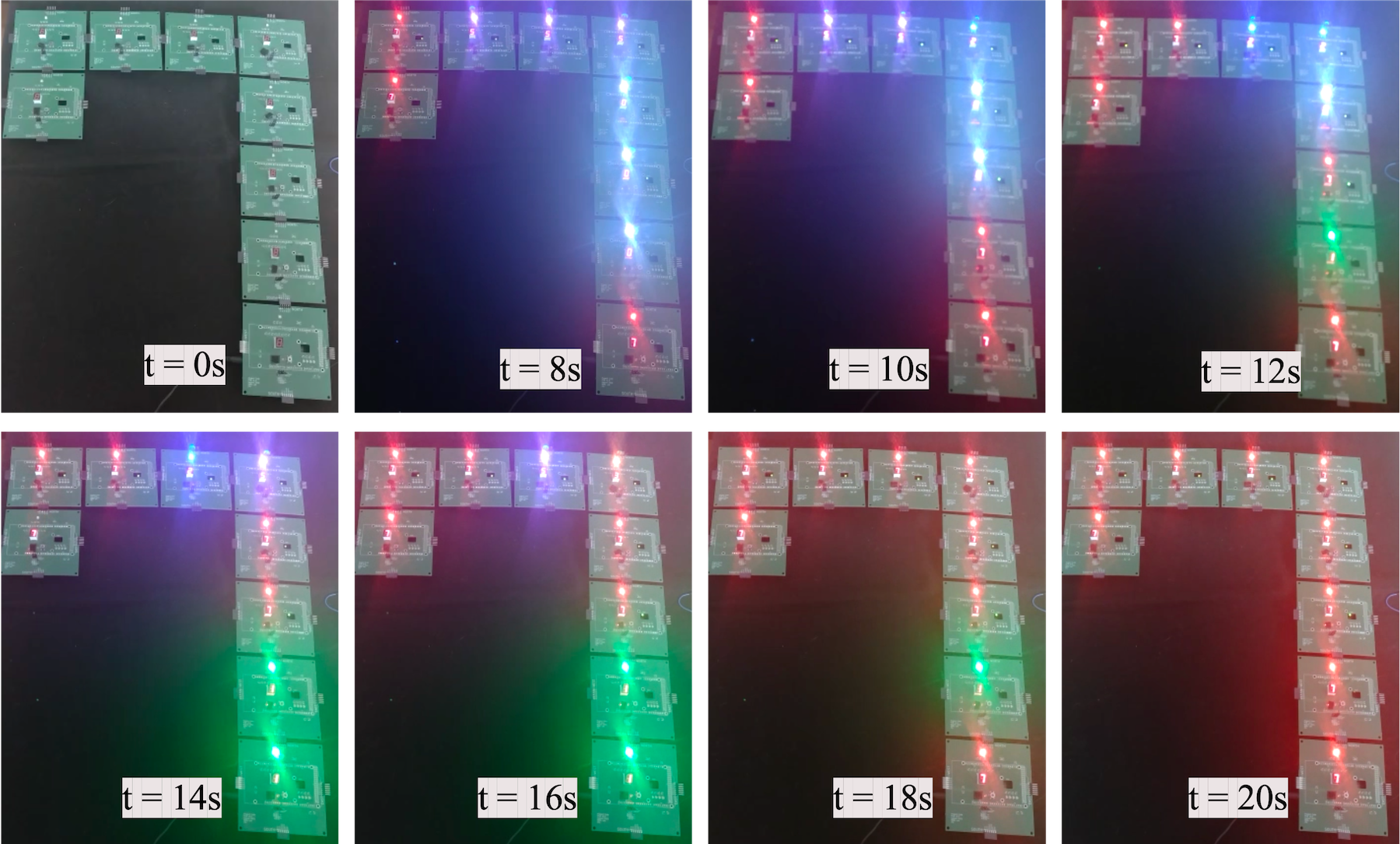}
         \caption{Results from hardware set-up of ``7".}
     \end{subfigure}
     \hfill
     \begin{subfigure}[b]{0.3\textwidth}
         \centering
         \includegraphics[width=\textwidth]{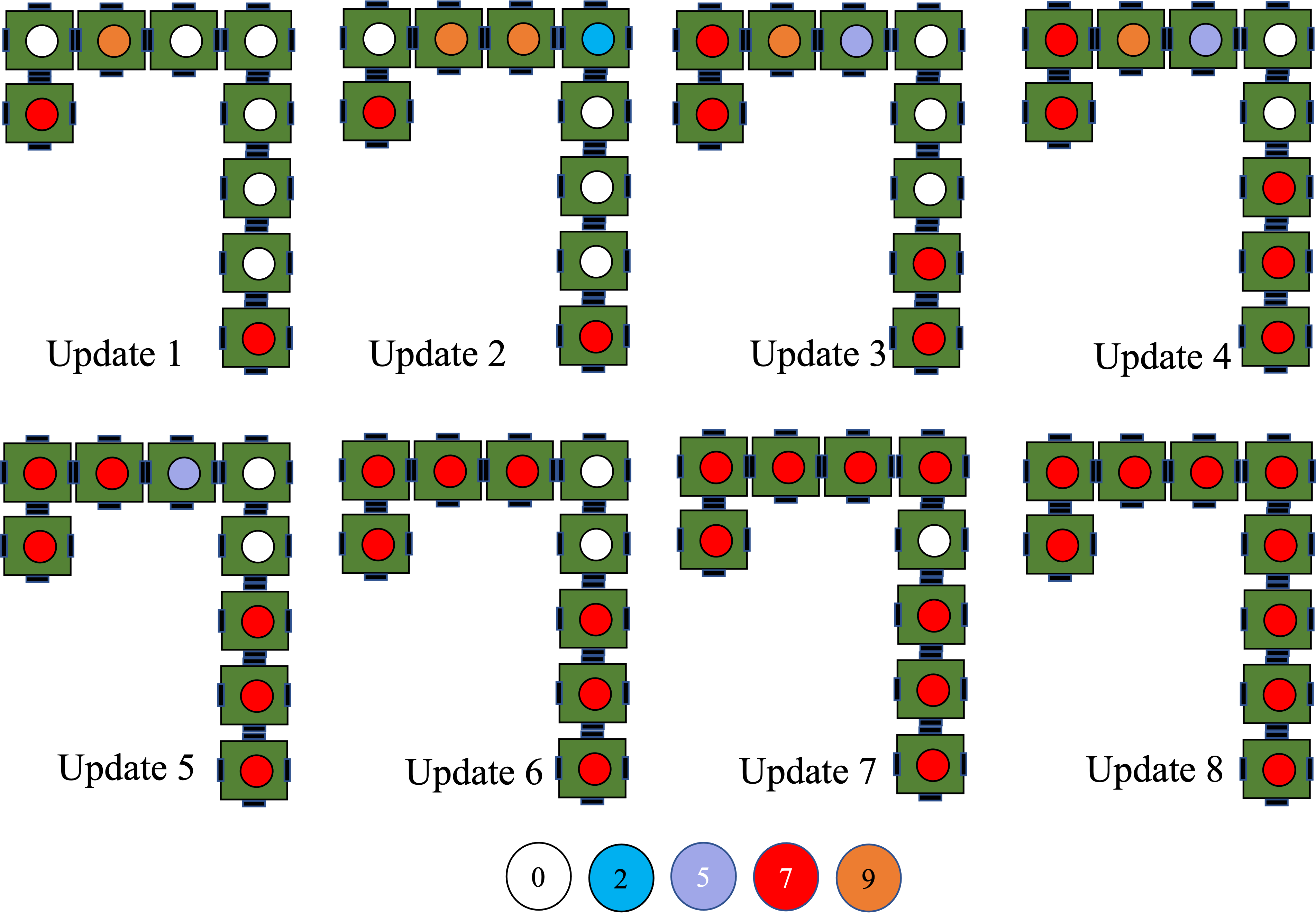}
         \caption{Results from the simulation set up of ``7".}
     \end{subfigure}
    \hfill
     \begin{subfigure}[b]{0.3\textwidth}
         \centering
         \includegraphics[width=\textwidth]{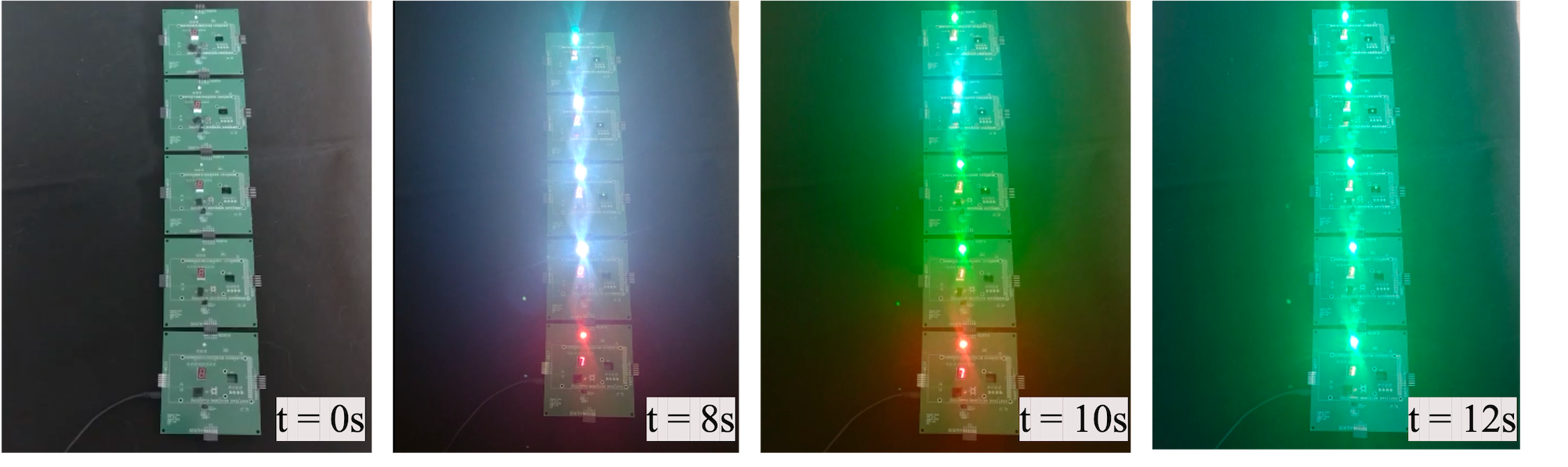}
         \caption{Results from hardware set-up of``1".}
     \end{subfigure}
     \hfill
     \begin{subfigure}[b]{0.3\textwidth}
         \centering
         \includegraphics[width=\textwidth]{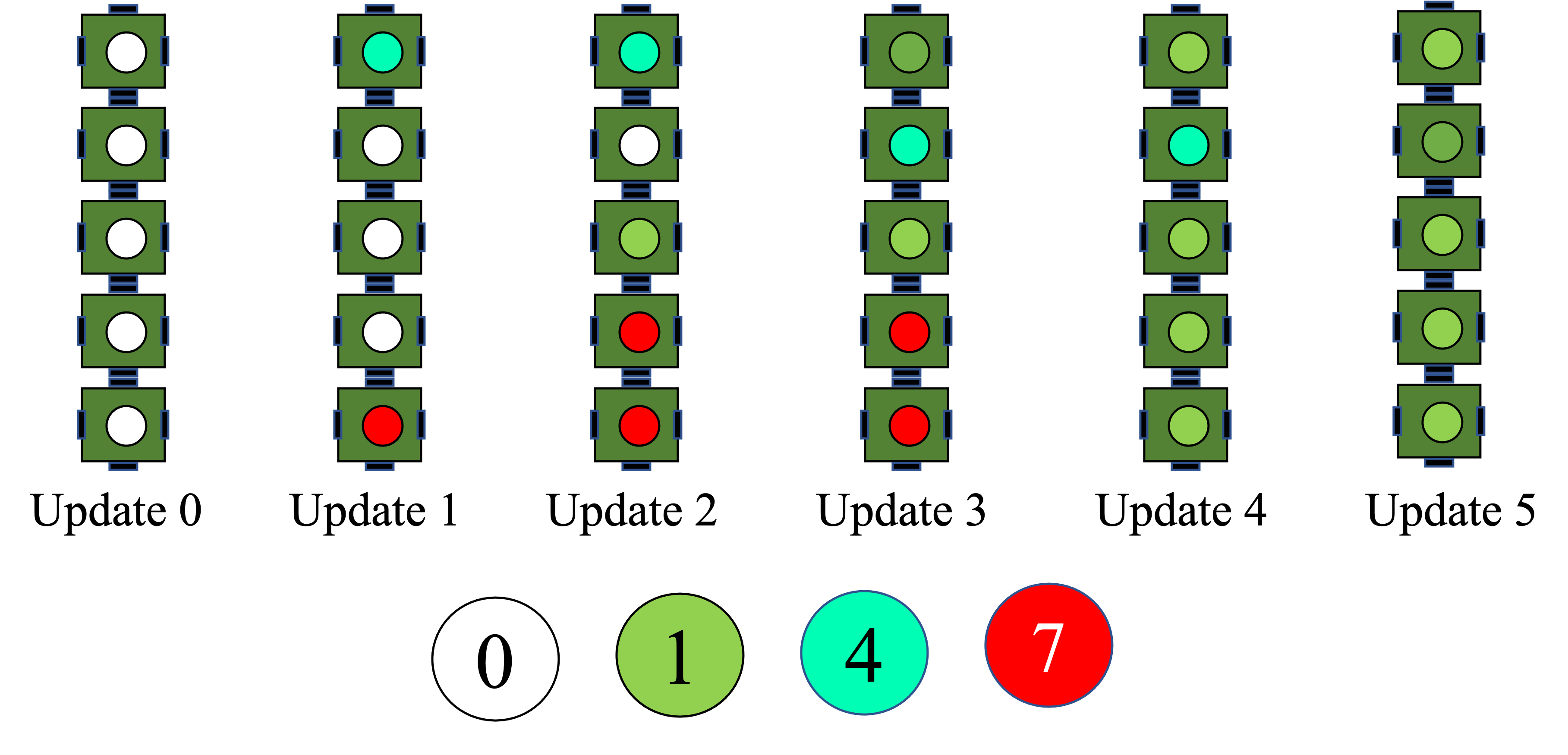}
         \caption{Results from the simulation set up of ``1".}
     \end{subfigure}
     \caption{Comparison of the results from simulation and hardware NCA for the numbers 4, 7, and 1.}
     \label{1_comparison}
     
\end{figure}

The first shape explored in detail was ``4".  The hardware system is able to successfully carry out the neural cellular automata process and all the cells correctly predict that the overall shape formed is the number 4 (Figure~\ref{1_comparison}a). This is indicated by both the blue LED light, and the number on the seven segment display. In Figure \ref{1_comparison}a we also show the approximate time taken for each of the updates (in this case 8 seconds before the first update, then the following updates take approx 2 seconds). This is somewhat arbitrary however, as the time between updates can be set within the firmware.

For comparison, the results from a simulated cellular automata are shown in Figure~\ref{1_comparison}b. Naturally, there are differences between the individual updates. This is due to the random update nature of both the hardware and the simulated version. However, importantly, both converge on the correct solution in approximately the same number of updates (i.e.\ 6 in hardware, 5 in this simulated example).

Note that in the simulated version, cells that have not yet had their first update are reported as a predicted ``0" value, whereas in hardware the remain unreported (i.e.\ no light turned on, and no value shown on the seven segment display.) Figure \ref{1_comparison}c-f show the hardware and simulation results for shapes 1 and 7.

These results demonstrate that the system designed and implemented in simulation can successfully be transferred to hardware. %Although only 3 different numbers/shapes are shown here, as previously stated, 
The hardware transformation is successful for all the shapes the neural network was trained on (i.e.\ 0 -- 9).

\subsection{Inherent Robustness}

Furthermore, we carried out experiments in both simulation and hardware to demonstrate inherent robustness present as a results of this type of self-classification. In particular, we test whether the neural cellular automata is able to determine the correct class of shape (i.e.\ the correct number) if the overall scale of the number was changed. Note that the neural network is  not trained explicitly for these different scales; it is only  trained on the original 4$\times$5 grid shapes. Successfully   classifying at other scales could point towards some inherent robustness in the system.

\begin{figure}[htbp!]
     \centering
     \begin{subfigure}[b]{0.45\textwidth}
         \centering
         \includegraphics[width=\textwidth]{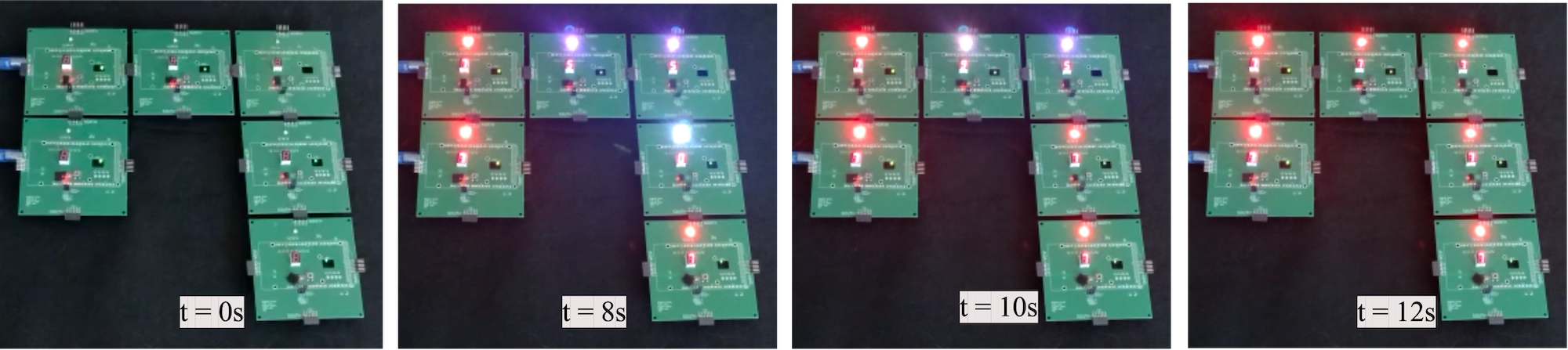}
         \caption{Hardware of smaller version of ``7".}
     \end{subfigure}
     \hfill
     \begin{subfigure}[b]{0.45\textwidth}
         \centering
         \includegraphics[width=\textwidth]{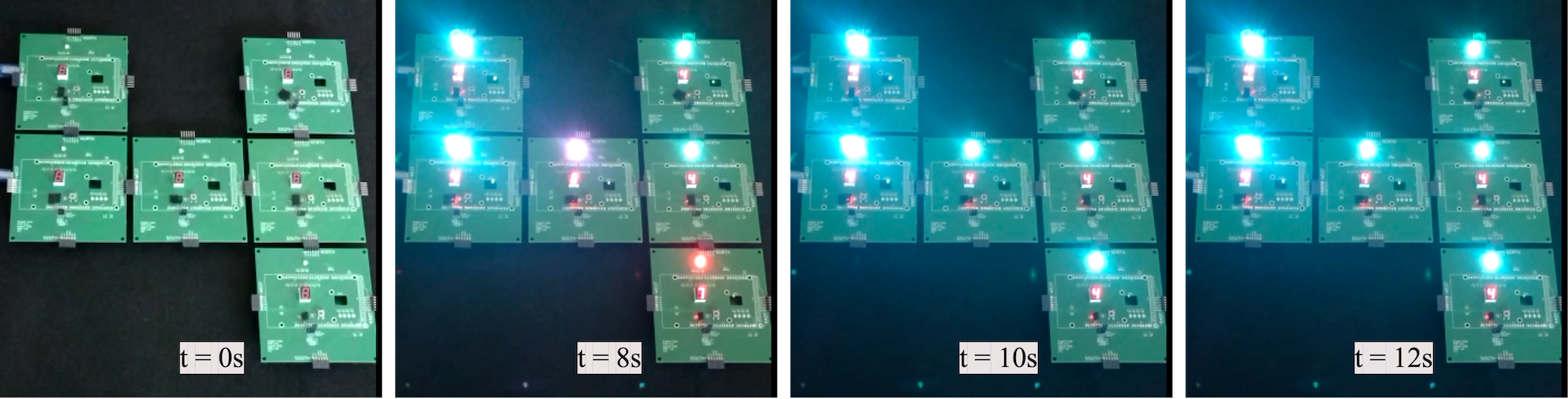}
         \caption{Hardware of smaller version of ``4".}
     \end{subfigure}
     \hfill
     \begin{subfigure}[b]{0.45\textwidth}
         \centering
         \includegraphics[width=\textwidth]{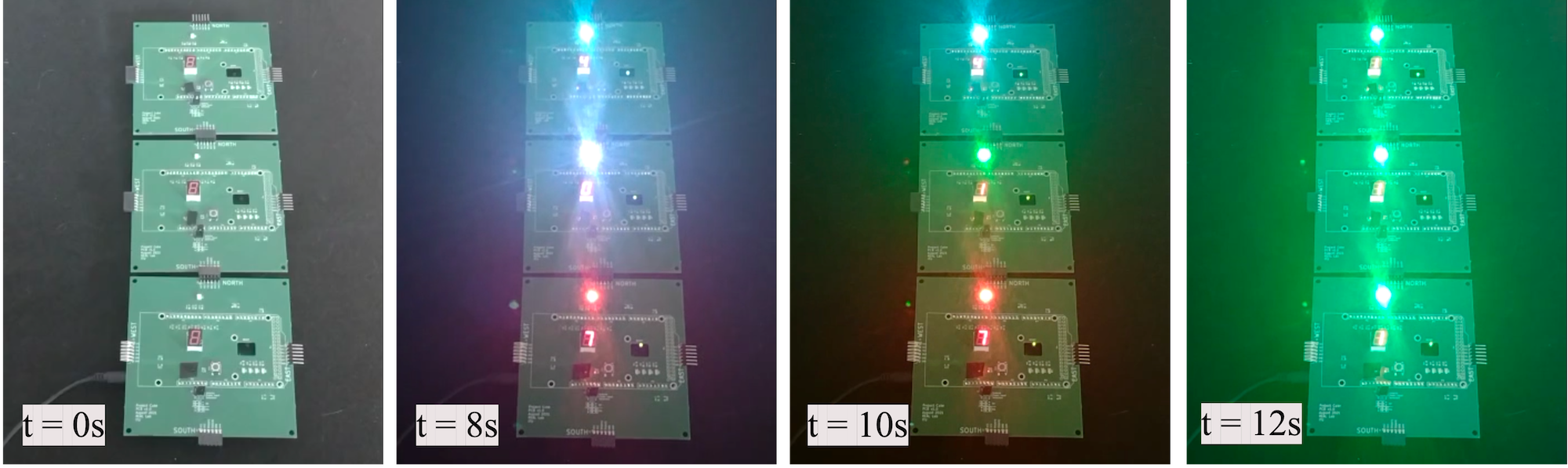}
         \caption{Hardware of smaller version of ``1".}
     \end{subfigure}
     \caption{Inherent robustness to scale changes. Shown are  examples of smaller versions of the numbers, for which the neural network was not trained for, yet the cells are  capable of agreeing on the correct shape.}
     \label{mini}
\end{figure}

As only 20 hardware tiles are manufactured so far, we tested only scaled down robustness in hardware (i.e.\  can the NCA  successfully classify the same shapes with fewer cells, in this case on a 3$\times$3 grid)  Furthermore, we only tested on a subset of scaled-down shapes -- numbers such as 2, 3, 5, 6 and 9 would lose their overall shape when scaled to this smaller grid size. %Therefore, in hardware we tested scaled down versions on 0, 1, 4, 7 and 8. 
For each of these scaled down numbers/shapes the neural cellular automata was able to classify successfully (see Figure~\ref{mini} for the 1, 4 and 7 examples) despite not being trained for them.

We also observed that in these scaled down scenarios, the amount of updates taken for the cellular automata to determine and agree on the shape classification was much smaller than for the original sizes. This is not unexpected as the amount of required communication between all the cells is also much less. 

\begin{figure}[t!]
     \centering
     \begin{subfigure}[b]{0.45\textwidth}
         \centering
         \includegraphics[width=\textwidth]{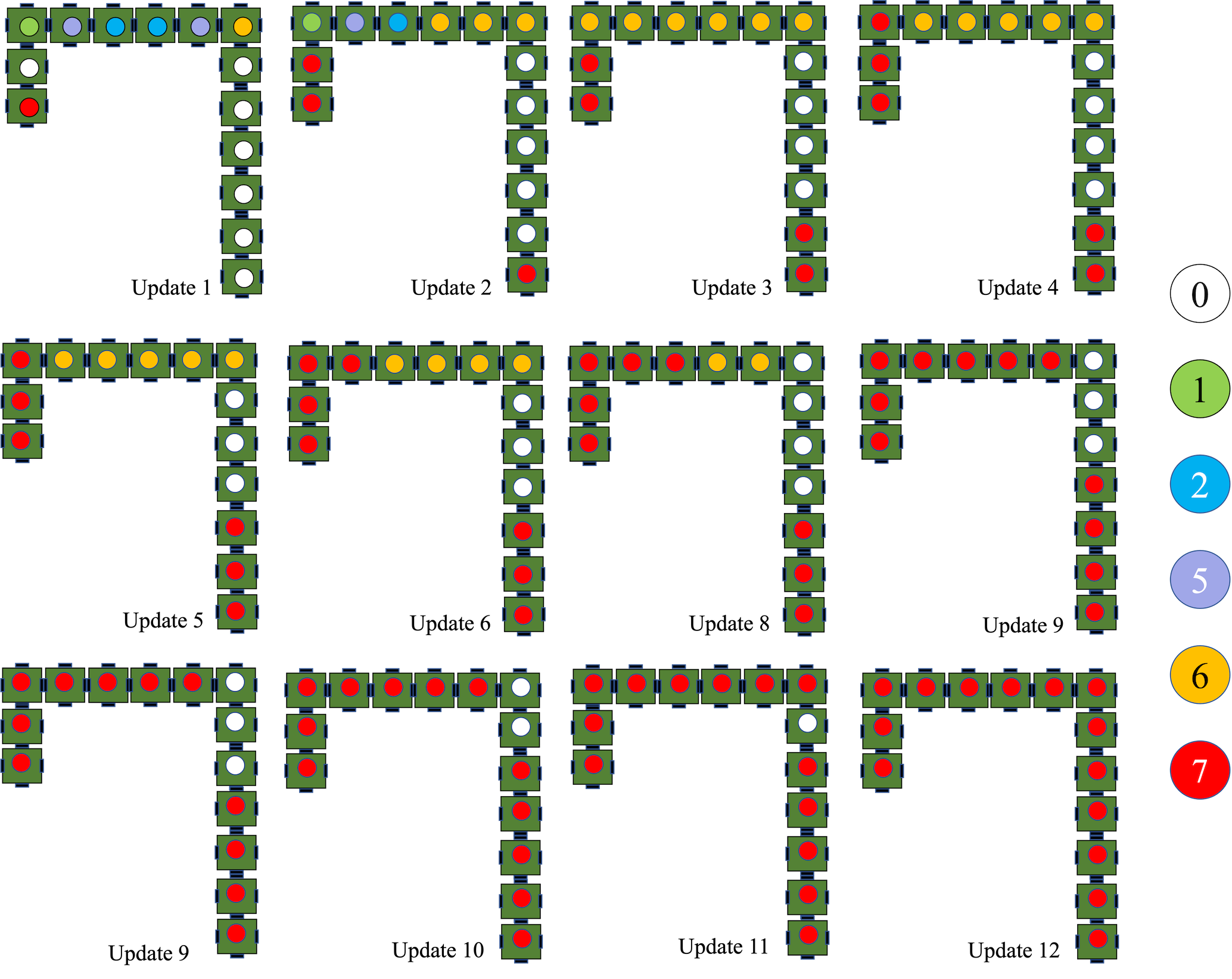}
         \caption{Simulation of scaled up ``7".}
     \end{subfigure}
     \hfill
     \begin{subfigure}[b]{0.45\textwidth}
         \centering
         \includegraphics[width=\textwidth]{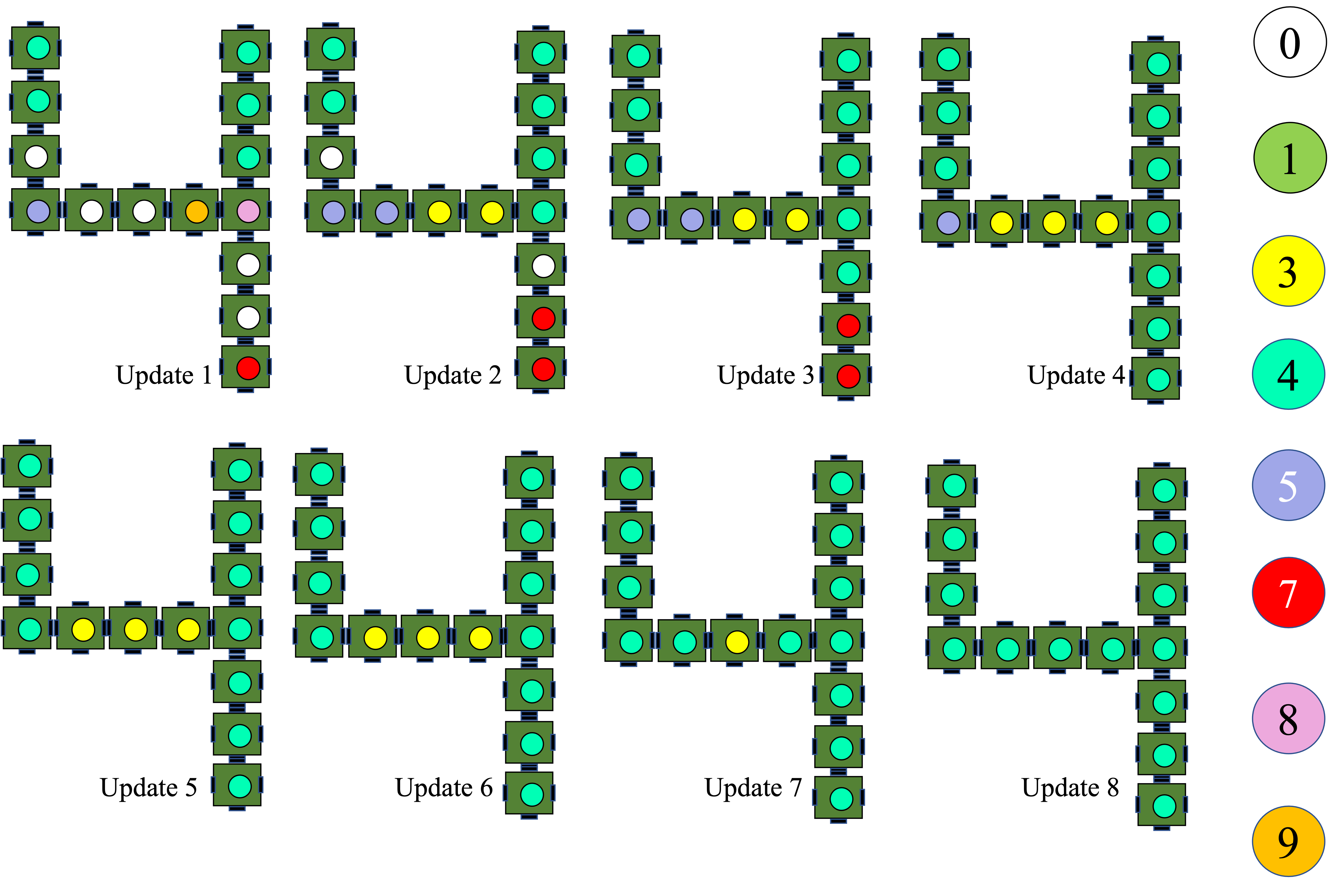}
         \caption{Simulation of scaled up ``4".}
     \end{subfigure}
     \hfill
     \begin{subfigure}[b]{0.45\textwidth}
         \centering
         \includegraphics[width=\textwidth]{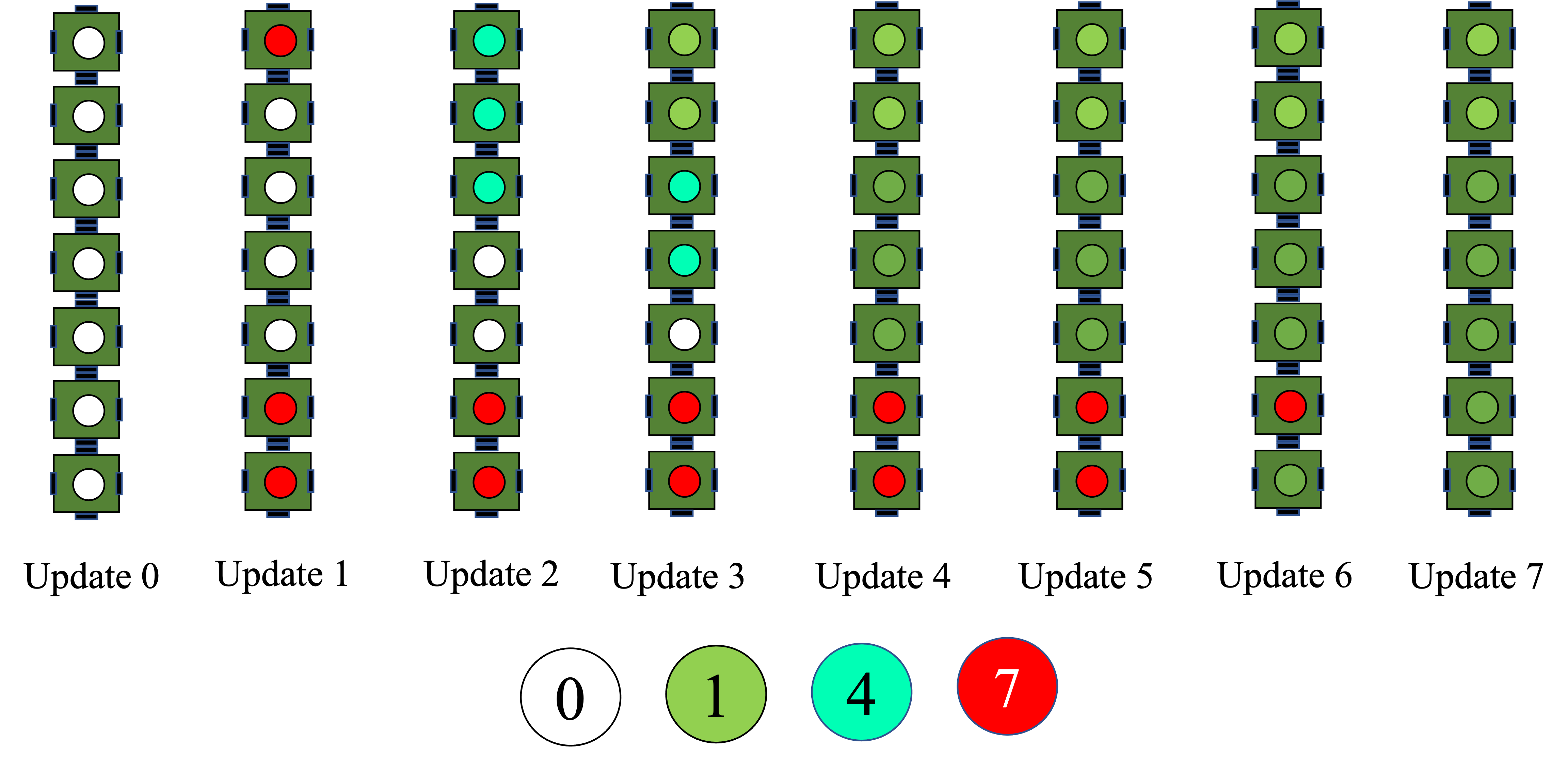}
         \caption{Simulation of scaled up ``1".}
     \end{subfigure}
     \caption{Scaled up versions of the trained numbers. Note that the neural network was not trained on these larger versions but the cells are still able to agree on the correct overall shape due to the inherent robustness in the system.}
     \label{scaled_up}
\end{figure}

We also tested whether this inherent robustness was present in scaled up versions of the original numbers/shapes, this time on a 6$\times$7 size grid and given the limited number of hardware tiles available only in simulation. All 9 scaled up versions of the shapes/numbers were once again successfully classified, with Figure~\ref{scaled_up} showing these results for 1, 4, and 7. In the future we also aim to test this scaled up version in hardware, which we believe should also work well, given the successful transfer from simulation to hardware demonstrated in  the previous experiments.  
%confident of success in the hardware domain as well. 

Note that in these scaled up version, the number of updates required for all cells to successfully self-classify is again larger than their original counter parts. We hypothesise that this is due to potentially important information having to travel longer distance now. 
%Given that there are now more overall cells to communicate with this is not surprising. 
For this grid size, the cellular automata is able to successfully classify every shape  within the 30 update limit enforced during training. However, it is our expectation that if the grid size was increased further, this may not be the case. Therefore, if an even larger scaled up version was required, the 30 update limit during training may have to be increased.

\section{DISCUSSION AND CURRENT LIMITATIONS}

The work presented here takes initial steps towards hardware systems that can determine the class of shape they belong to, only through the local communication of their components.
%We consider the work presented here as an important first step towards self classifying machines and robots, i.e., those with the capability through local communication to determine the class of shape they below to. 
In this section we discuss the limitations of the current research and future directions for improvements.

Firstly, the amount of tiles manufactured was minimal. Results from simulation show that the system is capable of scaling up the number of cells in each shape without loss of performance and with minimal increase in data stored on each tile (some increase in neural network size may be required). We are therefore confident that the results should also scale in hardware. Thus, one exciting area of further research is to manufacture more 2D tiles and explore the self classification of more complex shapes. Furthermore, previous work such as that by Rubenstein et al.~\cite{rubenstein2014programmable} highlighted the necessity of studying natural phenomena at sufficient complexity -- a strong motivation for this area of future work. We anticipate that a small redesign may be required as currently the electronics are not optimised for large power transfer through the connectors and low power consumption.%large power consumption 

Furthermore, few machines and robots are designed and built to operate purely in 2D as this can cause problems when interacting with the real 3D world. Previously published simulated works on neural cellular automata, such as that by Sudhakaran et al.~\cite{sudhakaran2021growing} (and also \cite{horibe2021regenerating}) show success in 3D domains. In addition to scaling up the number of cells and therefore the complexity of the shapes the neural cellular automata are trained on, we also aim to investigate performance in 3D. This would be another crucial step towards self-classifying and self-modelling materials.% classifying, complex robots. 

A current limitation with our design is that whenever a shape change occurs, all cells must be reset and re-start the classification process from the beginning. Therefore, cells cannot be added or removed \textit{on the fly} and the system cannot adapt and update its classification automatically. One of our motivations for future research is to allow machines to detect damage and automatically adjust their behaviour to compensate.  With the current firmware design this is not possible, however, on-the-fly adaption has been proved to work in simulation \cite{randazzo2020self} and we expect that with a few adjustments it could also be achieved in hardware.

Finally, we hope to expand the capabilities of the current system to interact with its environment 
%by enabling further artificial intelligence and
through the addition of sensors and actuators. In this context, combining the ability to self-classify with the ability to consider outcomes of multiple future actions through self-modeling \cite{chen2022fully,bongard2006resilient} could further increase the potential application areas of these robotic systems.

\section{CONCLUSION}

We presented a system in which modular tile-based structures can self classify their own simple shapes based on neural cellular automata and local-only communication. Due to its ability to self-classify, the system shows some inherent robustness without being trained for it, such as recognizing smaller version of the shapes it was trained on. While currently only a proof-of-concept, the approach opens up interesting future research direction such as shape recognition of large-scale 3D structures, or self-modelling materials that can automatically detect damage. 

\section*{Acknowledgements}
This work was supported by a Sapere Aude: DFF-Starting
Grant (9063-00046B) and DFF-Research Project1 grant (9131- 00042B).

\bibliography{reference}

\end{document}